\documentclass[11pt]{article}

\usepackage[utf8]{inputenc}
\usepackage[T1]{fontenc}
\usepackage{lmodern}
\usepackage[margin=1in]{geometry}
\usepackage{graphicx}
\usepackage{multirow}
\usepackage{amsmath,amssymb,amsfonts}
\usepackage{amsthm}
\usepackage{mathrsfs}
\usepackage{xcolor}
\usepackage{textcomp}
\usepackage{booktabs}
\usepackage{array}
\usepackage{float}
\usepackage{placeins}
\usepackage{caption}
\usepackage{natbib}
\setcitestyle{citesep={,}}
\usepackage{url}
\usepackage[hidelinks]{hyperref}

\graphicspath{{figures/}{./}{../figures/}}

\theoremstyle{plain}

\theoremstyle{definition}

\newcommand{\dratio}{\ensuremath{d_{\mathrm{wrong}-\mathrm{right}}}}
\newcommand{\bcorrect}{\ensuremath{\beta_{\mathrm{correct}}}}
\newcolumntype{L}[1]{>{\raggedright\arraybackslash}p{#1}}
\providecommand{\botrule}{\bottomrule}
\newcommand{\bmhead}[1]{\paragraph*{#1.}}
\providecommand{\orcidlink}[1]{\href{https://orcid.org/#1}{\nolinkurl{#1}}}

\title{Humans Disengage, Reasoning Models Persist\\
Separating Difficulty Registration from Deliberation Allocation}

\author{Han-yu Wang\\
The University of Hong Kong\\
\href{mailto:henry.why@connect.hku.hk}{\texttt{henry.why@connect.hku.hk}}}
\date{}

\begin{document}
\maketitle
\begin{abstract}
\noindent
Large reasoning models (LRMs) tend to produce longer reasoning traces on problems that also take humans longer. This correspondence leaves open how the systems distribute further work on those problems. We distinguish \textit{difficulty registration}, sensitivity to differences in problem difficulty, from \textit{deliberation allocation}, the distribution of further work once difficulty is encountered. We examine both in item-matched data from three reasoning tasks. In visual abstraction (H-ARC), model trace length follows the human ordering of problems by duration. After item identity is controlled, successful human attempts last longer than failed attempts, while failed LRM attempts have longer traces than successful ones in the pooled model analysis. The estimated slopes follow the same pattern in intuitive reasoning (INTUIT). In relational reasoning (Cortes), successful attempts are longer in separate human and model analyses, while a joint fit on shared items finds a human-LRM difference. Longer human attempts include more grid actions. At comparable lengths, failed LRM traces contain more hedging on H-ARC and more repetition on Cortes. A resource-rational account relates these patterns to what further work is expected to achieve and how that progress is valued. The results identify a difference in the allocation of continued work that cross-problem duration alignment alone leaves undetected.
\end{abstract}


\section{Introduction}
\label{sec:intro}

Research on deliberation asks how a cognitive system decides whether further thought is worth its cost. Sequential-sampling models distinguish an evolving evidence state from the stopping rule applied to it \citep{ratcliff2008diffusion,bogacz2006,heitz2014sat}. Metacognitive theories distinguish monitoring of confidence or uncertainty from the control of continued effort \citep{nelson_narens_1990,yeung2012metacog,nelson1988laborinvain}. Resource-rational metareasoning asks whether another computation promises enough improvement to justify its opportunity cost \citep{simon1956,russell1991,lieder2020bbs,gershman2015computational,callaway2022rational}. Each account distinguishes information about the current state from the rule that determines whether work continues, stops, or changes direction.

Further thought can remain worthwhile on a difficult problem if continued search still seems promising. The same difficulty can also prompt withdrawal when progress seems unlikely, or a change of approach when the current route loses value. We use \textit{difficulty registration} for sensitivity to differences in problem difficulty, expressed in observable deliberation, and \textit{deliberation allocation} for how an agent distributes further work once difficulty is encountered. Two agents can register a similar ordering of difficult problems and differ in when they persist.

We examine these relations through the duration of individual attempts. Across problems, we ask whether the same items elicit more deliberation from each system. Among attempts on the same problem, we ask how duration differs between those that eventually succeed and those that fail.

Large reasoning models (LRMs) make this comparison possible because they generate intermediate reasoning traces before a final answer \citep{wei2022cot,kojima2022zero,snell2024testtime}. Such behavioural signatures are increasingly used to study their relation to human cognition \citep{binz2023cognitive}. \citet{devarda2025} reported that LRM reasoning-token length tracks human reaction time across seven reasoning tasks. Problems that take humans longer also tend to elicit longer model traces. Although seconds and tokens measure different activities, their cross-problem correlation reveals a correspondence in the relative duration devoted to particular items. The question is how far this correspondence extends to continued work once difficulty has been encountered.

Two agents can share a cross-problem ordering even if one ends some difficult attempts quickly and the other produces long attempts that still end in error. Both may spend longer on difficult problems overall. A correlation across items leaves this difference unresolved because item difficulty and persistence contribute to the same duration measure. Comparing successful and failed attempts after item identity is held fixed separates the shared difficulty gradient from this further source of variation.

Recent debate over reasoning traces has made the limit of cross-problem correlation more salient. Behavioural correlation alone does not establish mechanistic similarity \citep{dujmovic2026commentary}. Increasing reasoning effort also produced little change in accuracy in five of six tasks examined by \citet{vankov2026commentary}. Intermediate tokens may serve as performative scaffolding and need not directly record cognitive cost \citep{hu2026commentary}. Replies defend the cross-problem regularity while leaving its algorithmic explanation open \citep{devarda2026replyduj,devarda2025reply}. This leaves a further behavioural question. Do humans and LRMs concentrate continued deliberation in the same kinds of trajectories when they encounter difficult problems?

We examine this question in three reasoning paradigms. H-ARC provides the primary test in visual abstraction, INTUIT extends the comparison to intuitive reasoning, and Cortes tests binary relational reasoning. The sample contains six open-weight thinking LRMs and DeepSeek-V3 as a non-thinking baseline. We compare the duration of successful and failed trials before estimating human and LRM outcome slopes with item fixed effects. Human grid actions and model trace content provide further evidence about longer attempts. The three paradigms allow us to test whether the human-LRM difference recurs across tasks and whether it always takes the form of opposite outcome slopes.

\section{Methods}
\label{sec:methods}
\label{sec:diagnostic}

\paragraph{Data and deliberation measures.}
This study reanalyses publicly released, de-identified human and LRM
data from the \citet{devarda2025} release and the underlying behavioural
corpora \citep{legris2024,prunty2025intuit,cortes2021}. No new
human-subjects data were collected. Table~\ref{tab:datasets} summarises
the paradigms. H-ARC \citep{legris2024,chollet2019arc} tests visual
abstraction, INTUIT tests intuitive reasoning, and Cortes tests binary
relational reasoning. Arithmetic is excluded because all thinking LRMs
are at ceiling.

The thinking-LRM sample consists of models that expose an intermediate
reasoning trace. It includes DeepSeek-R1 \citep{deepseek2025},
Qwen-QwQ-32B \citep{qwen2025} and Qwen3-235B-Thinking \citep{qwen3thinking2025},
GLM-4.5-Air-FP8 \citep{zhipu2025}, and gpt-oss-20b and gpt-oss-120b
\citep{openai2025}. DeepSeek-V3 \citep{deepseekv3_2024} is the
non-thinking baseline. On H-ARC, DeepSeek-R1, GLM-4.5-Air-FP8,
gpt-oss-120b, and Qwen-QwQ-32B each contribute at least $298$
observations. gpt-oss-20b has $n=119$ parseable trials. Qwen3-235B-Thinking has $n=43$ trials with four errors. These two H-ARC cells are
reported descriptively. All six thinking LRMs are included in the
INTUIT and Cortes analyses.

For thinking LRMs, deliberation $t$ is the
\texttt{reasoning\_token\_length} field in the released dataset. It
records the number of intermediate-trace tokens generated before the
final answer using each model's native tokenizer. For DeepSeek-V3,
we use \texttt{total\_output\_tokens}. These
quantities are treated as behavioural measures of generated duration
under nominally greedy decoding. Their relation to latent computation
is a subject of ongoing debate
\citep{stechly2025,samineni2025,kambhampati2025anthropomorph}. Cohen's $d$ standardizes each agent's descriptive outcome contrast, while
the regressions use logged duration. Both analyses concern relative
changes within a system instead of raw comparisons between seconds and tokens.
Following \citet{devarda2025}'s analysis script, we retain one
(item, model) row from the released dataset and keep the first attempt.
For cross-item duration correlations, model token counts are matched to item-level human reaction times irrespective of whether the model's final answer can be scored. Analyses involving correctness use only scored answers.
INTUIT traces are restricted in the public release, so trace-content
analyses use H-ARC and Cortes. SI-3 summarises provenance and scoring. The analysis code and complete
per-model attempted, parseable, correct, and wrong counts are in the
OSF deposit.

\begin{table}[!htbp]
\centering
\small
\begin{tabular}{@{}L{0.11\textwidth}L{0.18\textwidth}L{0.22\textwidth}L{0.39\textwidth}@{}}
\toprule
Paradigm & Cognitive domain & Human data & LRM data \\
\midrule
H-ARC & Visual abstraction & $400$ items, $4{,}091$ trials & $5$ thinking LRMs in the d-ratio summary, $6$ in the per-LRM item-FE regressions\textsuperscript{a} \\
INTUIT & Intuitive reasoning & $144$ items, $1{,}536$ trials & $6$ thinking LRMs with both outcomes \\
Cortes & Relational reasoning & $86$ items, $20{,}052$ trials & $6$ thinking LRMs ($5$--$8$ wrong trials each) \\
Arithmetic & Parametric arithmetic & $168$ items, $1{,}680$ trials & all LRMs at ceiling (excluded) \\
\botrule
\end{tabular}
\caption{Datasets analysed. \textsuperscript{a}\,gpt-oss-20b has $n=119$
parseable trials, and Qwen3-235B-Thinking has $n=43$ matched items
with four wrong trials. Full dataset and scoring detail are in the
OSF deposit.}
\label{tab:datasets}
\end{table}

\paragraph{Outcome-conditioned allocation and difficulty control.}
For agent $A$, paradigm $P$, and trial $i$, let $t_i>0$ denote
deliberation and $c_i\in\{0,1\}$ denote correctness. The per-agent
d-ratio is Cohen's $d$ on log deliberation using the pooled standard
deviation. Here $t_i$ is reaction time in seconds for humans,
\texttt{reasoning\_token\_length} for thinking LRMs, and total output
length for DeepSeek-V3.
\[
\dratio(A,P)=
\frac{\bar{\ell}_{\mathrm{wrong}}-\bar{\ell}_{\mathrm{right}}}
{s_{\mathrm{pooled}}},
\qquad \ell_i=\log t_i.
\]
Positive values indicate longer deliberation on trials that end in an
incorrect answer. We report a model d-ratio only when at least five correct and five incorrect trials are available.

The per-agent d-ratio combines between-item difficulty with differences
between successful and failed attempts. Harder items can take longer and
fail more often. Duration can also differ by outcome among attempts on
the same item. We separate these sources with item fixed effects. The
combined specification is
\[
\log t_i = \alpha + \beta c_i + \beta_{\mathrm{LRM}}\,
c_i\!\times\!\mathrm{LRM}_i + \delta_{\mathrm{agent}} +
\gamma_{\mathrm{item}} + \varepsilon_i.
\]
Here $\mathrm{LRM}_i$ equals one for model observations and zero for
human trials. The interaction $\beta_{\mathrm{LRM}}$ is the LRM
outcome-conditioned slope minus the human slope, with shared item effects. We also fit human-only and LRM-only item fixed-effects models,
single-LRM versions of the combined regression, the same specifications
on INTUIT and Cortes, and a within--between re-parameterisation \citep{mundlak1978}
using item-level mean correctness in place of item
dummies. Standard errors are cluster-robust by item throughout.

These coefficients relate duration to observed correctness. The human
slope compares trials on the same item across the sampled population,
whose participant identifiers are absent from the release. Each LRM
contributes one observed outcome per (item~$\times$~model) pair, so the
pooled LRM slope is identified by variation across models attempting the
same item. Repeated stochastic samples from one model are unavailable.
Separate subgroup fits give each group its own item effects. Further
specification details are in SI-2 and SI-5.

For H-ARC sensitivity analyses, we fit
$\log t \sim c_c + \mathrm{LRM} + c_c\!\times\!\mathrm{LRM}$ with
centred correctness, an LRM indicator, and an item random intercept. We
also fit crossed random effects and cluster-robust OLS specifications
clustered by agent and by item (Table~\ref{tab:harc-spec-si}). The item
fixed-effects estimate with item-clustered standard errors is the primary
difficulty-controlled estimand.

\paragraph{Trace-content analysis.}
For each LRM trial on H-ARC and Cortes, we compute four length-normalised
features from the released \texttt{reasoning\_trace}. These are
hedge markers per $1{,}000$ characters
(\texttt{wait}, \texttt{actually}, \texttt{hmm},
\texttt{let me reconsider}, \dots), self-correction markers
(\texttt{scratch that}, \texttt{ignore that}, \dots), word-level
$5$-gram repetition rate, and type--token ratio. For each feature we fit
\[
\text{feature}\sim\text{correct}+\log w+C(\text{agent})+C(\text{item}),
\]
where $w$ is trace length in words. Standard errors are cluster-robust
by item. Full marker lists and the parser are in the OSF deposit. SI-4
reports the treatment of multiple comparisons.

\section{Results}
\label{sec:results}

\begin{figure}[!htbp]
  \centering
  \includegraphics[width=\textwidth]{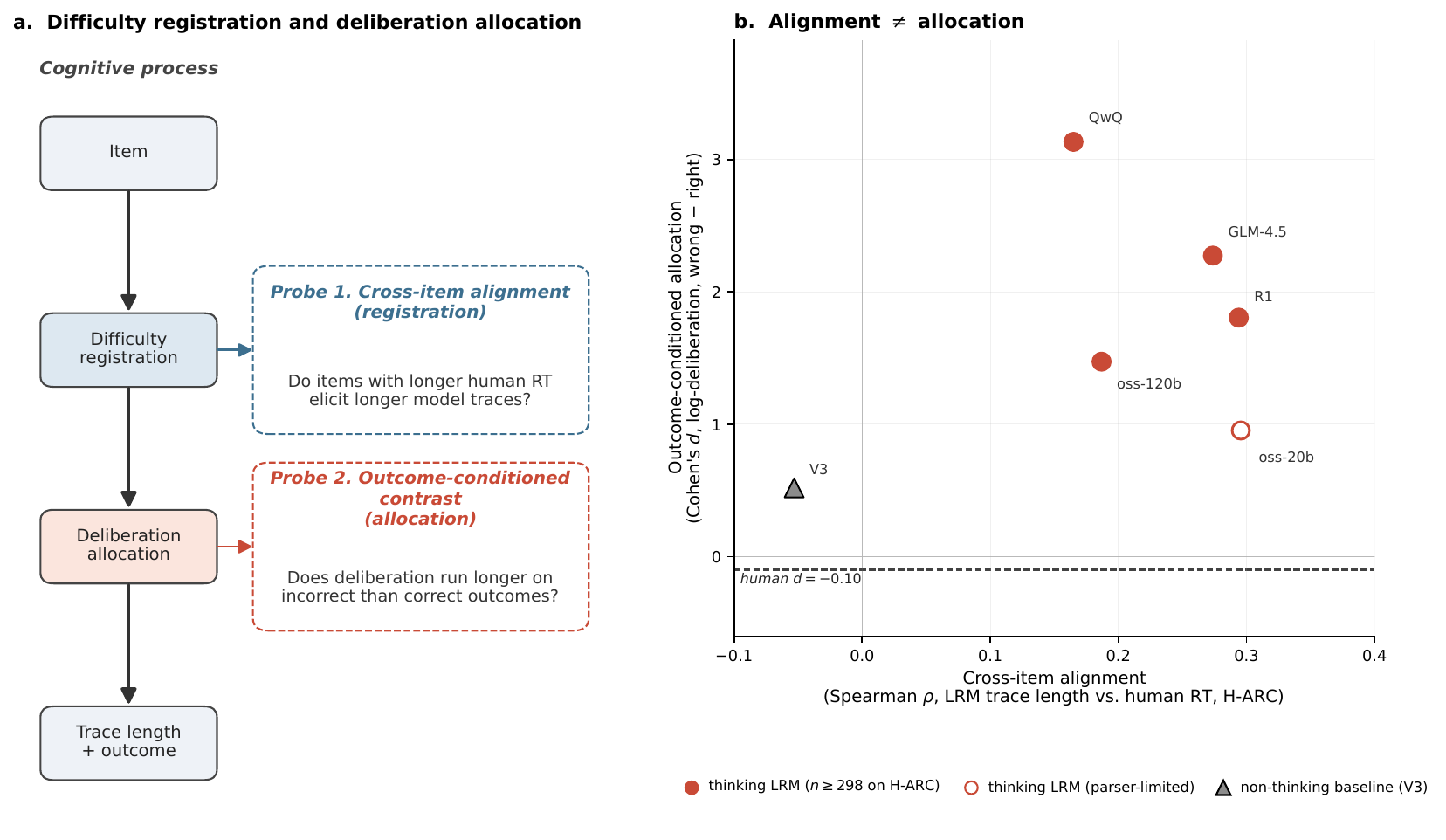}
  \caption{\textbf{Difficulty registration and deliberation allocation.}
  (a)~Difficulty registration describes how duration varies across items.
  Deliberation allocation describes how duration relates to outcome.
  (b)~H-ARC results. The horizontal axis is the cross-item Spearman
  $\rho$ between LRM reasoning-token length and human reaction time. The
  vertical axis is per-agent Cohen's $d$ on log deliberation (wrong $-$
  right). The open circle marks gpt-oss-20b. Its cross-item correlation
  uses $400$ duration pairs, while its d-ratio uses $119$ scored outputs. The dashed line marks the matched human $d=-0.10$, and the
  triangle marks non-thinking DeepSeek-V3.}
  \label{fig:diagnostic}
\end{figure}

Across the five thinking LRMs with reported d-ratios, trace length correlates positively with human reaction time across H-ARC items, with Spearman $\rho=0.16$--$0.30$ and $p<.001$. DeepSeek-V3 has $\rho=-0.05$ (Figure~\ref{fig:diagnostic}). Alongside this cross-item correspondence, all five thinking LRMs with estimable d-ratios have $d>0$, whereas the matched human sample has $d=-0.10$. Problems that elicit more deliberation from humans tend to elicit longer thinking-model traces, yet the systems differ in whether longer attempts end successfully.

\subsection{The allocation gap on H-ARC}
\label{sec:results-harc}

\begin{figure}[!htbp]
  \centering
  \includegraphics[width=\textwidth]{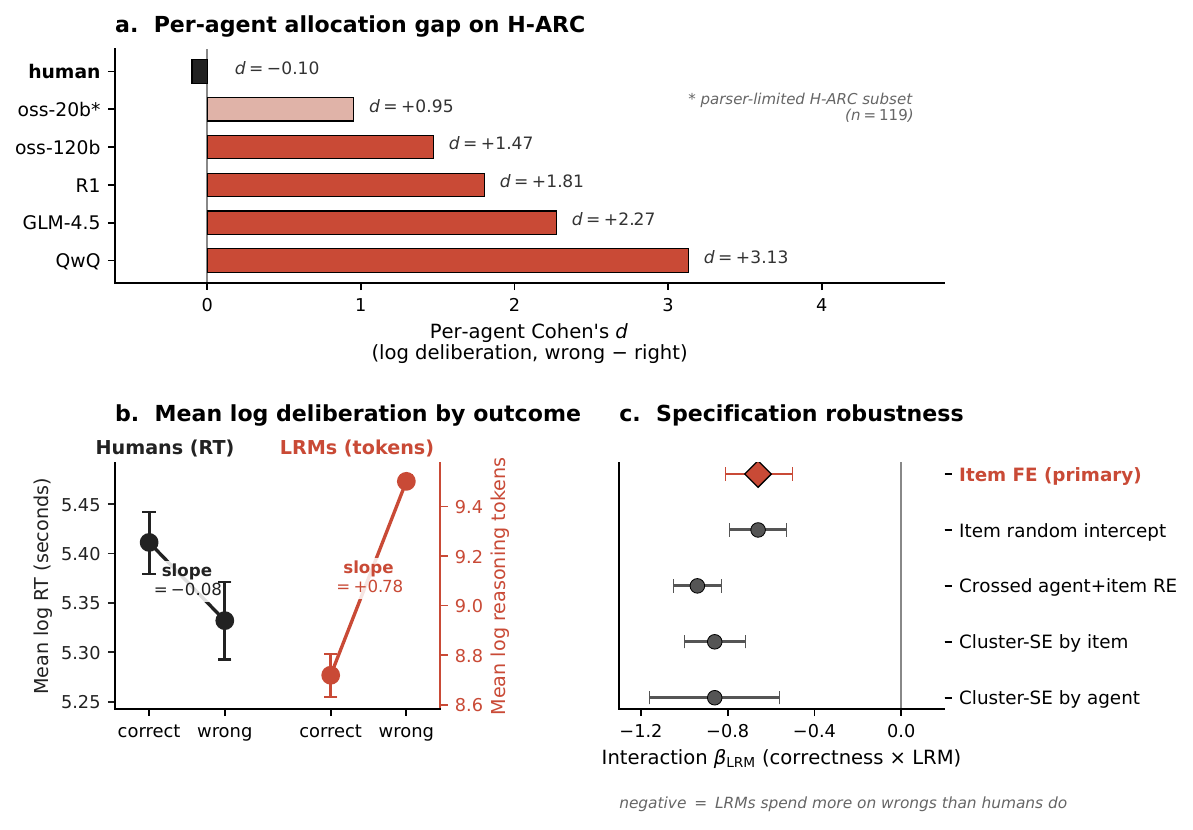}
  \caption{\textbf{The allocation gap on H-ARC.}
  (a)~Per-agent Cohen's $d$ on log deliberation (wrong $-$ right).
  The asterisk marks gpt-oss-20b, for which $119$ H-ARC outputs are
  parseable. Qwen3-235B-Thinking
  is omitted because only four H-ARC trials are wrong. Its item-FE
  interaction is reported descriptively in Table~\ref{tab:itemfe-perlrm-si}. (b)~Estimated
  marginal mean log deliberation by outcome with bootstrapped $95\%$
  CIs. Human RT and LRM reasoning tokens use separate y-axes. Their
  raw levels are not directly comparable. (c)~Interaction estimates
  across sensitivity specifications. The diamond is the primary
  item-fixed-effects estimand (Table~\ref{tab:harc-spec-si}).}
  \label{fig:harc-primary}
\end{figure}

H-ARC, the visual abstraction task, provides the primary inferential test and the largest matched model sample. Every thinking LRM with an estimable H-ARC contrast has higher mean log duration on wrong trials. Among the four model samples with at least $298$ observations, Cohen's $d$ ranges from $1.47$ to $3.13$. The matched human sample has $d=-0.10$ (Figure~\ref{fig:harc-primary}a and Table~\ref{tab:itemfe-perlrm-si}). The pooled descriptive means in Figure~\ref{fig:harc-primary}b point in the same directions. On wrong trials, mean human log reaction time is $0.08$ lower and mean LRM log trace length is $0.78$ higher than on right trials. The pooled means are plotted on separate axes. We test the human-LRM difference using item-controlled outcome slopes.

The cross-system difference remains after controlling item identity. The primary item fixed-effects model gives an agent-type $\times$ correctness interaction of $\beta_{\mathrm{LRM}}=-0.66$ log units ($p<.001$, $95\%$ CI $[-0.81,-0.50]$, $n=5{,}728$). In this joint model, successful human attempts last longer than failed attempts on the same item, while failed attempts have longer traces in the pooled LRM analysis. The interaction measures the difference between these slopes after the common item gradient has been removed.

Four sensitivity specifications give estimates from $-0.66$ to $-0.94$, with all $95\%$ CIs below zero (Figure~\ref{fig:harc-primary}c and Table~\ref{tab:harc-spec-si}). The within--between re-parameterisation \citep{mundlak1978} gives $\beta_{\mathrm{LRM}}=-0.79$ ($p<.001$). Within-item and between-item slopes differ at $p<.001$ in both agent groups (Table~\ref{tab:itemfe-mundlak-si}). LRM d-ratios are also positive in nearly every difficulty quartile (SI-7, Figure~\ref{fig:robustness}). Collapsing repeated human trials to item$\times$correctness cells gives $\beta_{\mathrm{LRM}}=-0.66$ with cell-size weighting and $-0.61$ without weighting (SI-2).

DeepSeek-V3 also produces longer outputs on errors despite lacking an explicit thinking trace. On Cortes it has $d=1.41$ ($95\%$ CI $[1.06,1.84]$). Its H-ARC estimate is $d=+0.52$ with a $95\%$ bootstrap CI of $[-0.22,+1.24]$, based on only $18$ correct trials ($4.5\%$ accuracy). INTUIT accuracy is too close to floor for an estimable contrast. On H-ARC, the estimated wrong-right duration contrast is positive despite a cross-item correlation of $\rho=-0.05$ with human reaction time. In the Cortes sample, error-associated output expansion is present in ordinary generation as well.

\subsection{Allocation across reasoning tasks}
\label{sec:results-generalisation}

\begin{figure}[!htbp]
  \centering
  \includegraphics[width=\textwidth]{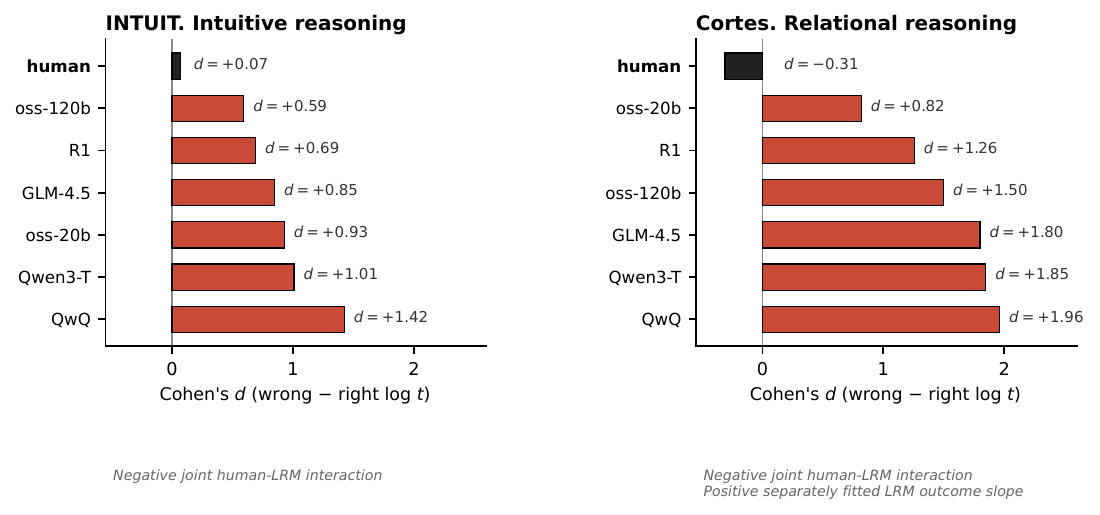}
  \caption{\textbf{Cross-paradigm allocation gap.} Per-agent Cohen's $d$ on log deliberation (wrong $-$ right)
  for the two non-saturated paradigms beyond H-ARC. INTUIT
  (intuitive reasoning) has estimated item-controlled human and pooled LRM slopes in the same directions as H-ARC. Cortes
  (binary relational reasoning) tests how the allocation contrast changes with the task. The combined
  agent-type interaction is negative in both paradigms. On Cortes,
  the separately fitted pooled LRM slope is positive. H-ARC is
  the primary analysis and is shown in Figure~\ref{fig:harc-primary}a.}
  \label{fig:paradigms}
\end{figure}

On INTUIT, every analysed thinking LRM has a positive d-ratio ($d=0.59$--$1.42$ across six models), compared with $d=+0.07$ for humans. The item-controlled interaction is $\beta_{\mathrm{LRM}}=-0.41$ ($p<.001$). The separately fitted human slope is positive ($+0.10$, $p=.087$) and the pooled LRM slope is negative ($-0.33$, $p=.013$). Their directions follow the H-ARC pattern, with the significant interaction providing the evidence for a cross-system difference (Figure~\ref{fig:paradigms}).

Cortes, the binary relational-reasoning task, shows a different relation between duration and outcome. Marginally, every thinking LRM has longer traces on wrong trials ($d=0.82$--$1.96$), compared with a human d-ratio of $-0.31$. When humans and LRMs are fitted separately with their own item effects, the pooled LRM correctness slope is $+0.31$ and the human slope is $+0.42$. Fitting the groups together on the $58$ shared items gives a human slope of $+0.47$ and an LRM slope of $-0.50$, producing a negative agent-type interaction of $\beta_{\mathrm{LRM}}=-0.97$ ($p<.001$).

The change in the LRM slope comes from changing the comparison. The human-only fit uses $86$ items, while the model and joint fits use $58$. The separate model fit estimates its item effects from model observations alone. The joint fit restricts both groups to shared items and estimates a common item effect from the much larger human sample together with the model observations. The interaction is negative in every single-LRM joint fit (Table~\ref{tab:perlrm-cross}). Each model, however, has only $5$--$8$ errors. Bootstrap intervals for the marginal d-ratio lie above zero for all six thinking LRMs (Figure~\ref{fig:cortes-ci}). Cortes thus retains a difference in the joint comparison without reproducing the opposite subgroup slopes found on H-ARC and INTUIT.

\subsection{Human engagement and the form of failed model traces}
\label{sec:results-mechanism}

For humans on H-ARC, the marginal d-ratio is close to zero at $d=-0.10$. The item-controlled correctness slope is $+0.24$ log units. The two quantities differ because within-item and between-item associations point in opposite directions. Within an item, solved trials are longer. Across items, easier problems are faster and more likely to be solved. At the item level, mean first-attempt reaction times yield $d=+0.60$ when items with released accuracy at or below $50\%$ are compared with those above it. The Mundlak decomposition gives a within-item slope of $+0.24$ and a between-item slope of $-0.54$ (Table~\ref{tab:itemfe-mundlak-si}).

\begin{figure}[!htbp]
  \centering
  \includegraphics[width=\textwidth]{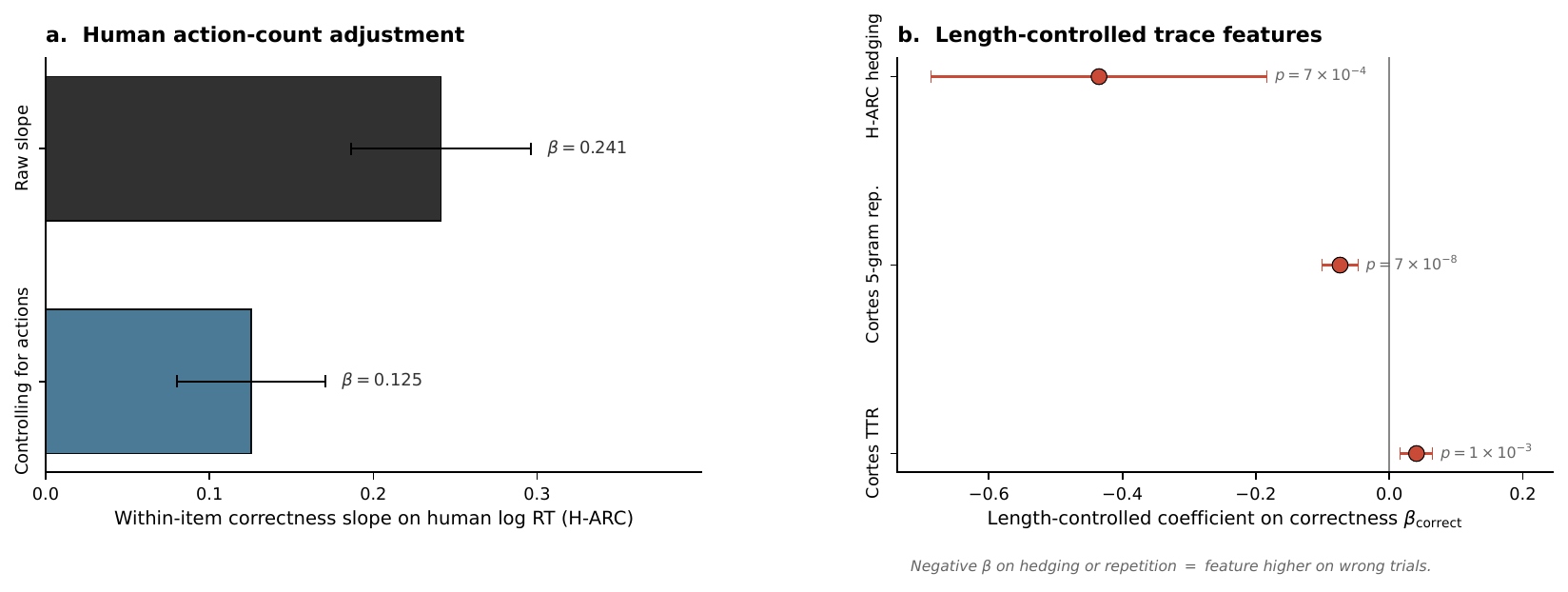}
  \caption{\textbf{Within-system behavioural probes.} (a)~Human action-count adjustment. The unadjusted
  item-controlled correctness slope on H-ARC human log RT is
  $\beta=+0.241$. Adding $\log(1+\text{actions})$ (the
  per-trial grid-action count) reduces it to $\beta=+0.125$, a
  $48\%$ reduction. (b)~LRM trace-content asymmetries.
  Length-controlled coefficients on correctness for the three
  emphasised feature$\times$paradigm cells of
  Table~\ref{tab:trace-content}. TTR denotes type--token ratio. Negative $\beta$ on hedging or
  repetition means the feature is \textit{higher} on wrong
  trials.}
  \label{fig:mechanism}
\end{figure}

Among wrong H-ARC trials, the fastest quartile has a median of $16$ grid edits and the slowest has $79$. Adding $\log(1+\text{actions})$ to the item-controlled regression reduces the correctness slope from $+0.241$ to $+0.125$ (Figure~\ref{fig:mechanism}a). Part of the extra time on successful attempts is thus associated with more grid activity. The action records also show that substantial work can end in failure. Short failures with few edits and long failures with many edits are consistent with selective persistence, in which some attempts end early and others sustain work without reaching a solution.

The estimated within-item correctness slope is larger on the hardest items ($\bcorrect{}^{\mathrm{human}}=+0.32$, $p<.001$) than on the easiest quartile ($+0.13$, $p=.054$). Across the $341$ items with both outcomes, greater reaction-time dispersion is associated with larger correctness slopes (Spearman $\rho=+0.27$, $p=5\times10^{-7}$). This last association is descriptive because both quantities are derived from the same reaction-time distribution.

After controlling trace length, item, and model, wrong H-ARC traces contain $0.43$ more hedge markers per $1{,}000$ characters than right traces ($\beta_{\mathrm{correct}}=-0.435$, $p=7\times10^{-4}$). The unadjusted wrong-right difference has the same direction in all five models with estimable H-ARC contrasts. On Cortes, wrong traces have a $7.4$ percentage-point higher $5$-gram repetition rate ($\beta=-0.074$, $p=7\times10^{-8}$) and a $4.1$ percentage-point lower type--token ratio ($\beta=+0.041$, $p=10^{-3}$). The unadjusted differences in both features have the same direction in all six models. H-ARC failures express more explicit hedging, while Cortes failures contain more repetition and less lexical diversity.

\section{Discussion}
\label{sec:discussion}

\subsection{From difficulty registration to control}
\label{sec:disc-reveal}

The distinction between registration and allocation matters most clearly on H-ARC. Problems that take humans longer also tend to elicit longer model traces. Once item identity is controlled, successful human attempts are longer and failed model attempts have longer traces in the pooled analysis. The estimated slopes on INTUIT point in the same directions. Cortes retains a difference in the joint comparison, although both subgroup slopes are positive when fitted separately.

Some human H-ARC failures involve few grid operations, while others follow substantial active work. Successful attempts on the same problem are longer on average, and action-count adjustment links part of this extra time to engagement. The selective-persistence interpretation accommodates both early endings and continued work that still fails. Across problems, however, easier items are solved more often and more quickly. These opposing relationships leave the marginal human d-ratio close to zero. The aggregate comparison can therefore suggest little difference between success and failure even when duration is systematically related to both item difficulty and the work sustained within an attempt.

Failed and successful model traces differ even at comparable length. H-ARC failures contain more explicit hedging, while Cortes failures contain more repetition. The extra length on failed attempts thus comes with a difference in the form of the generated work. On H-ARC the text more often signals hesitation, and on Cortes it returns more often to wording already used. These features help describe what models do during attempts that continue without reaching a correct answer.

DeepSeek-V3 produces longer outputs on Cortes errors despite having no explicit thinking trace. In this sample, error-associated expansion is not confined to models that expose a reasoning trace.

In metacognitive theories, confidence and uncertainty supply information about ongoing cognition, while control determines what the system does with it \citep{nelson_narens_1990,yeung2012metacog,nelson1988laborinvain}. Difficulty registration is broader than monitoring as used in these theories. It is defined behaviourally and can be measured without an explicit confidence report. Human confidence and model representations could differ substantially while producing a similar ordering of problems by duration. Allocation asks whether sensitivity to those problems also has comparable consequences for continued work. Human actions and model traces allow this comparison to include the work performed during deliberation.

Sequential-sampling models express the issue through the joint dependence of response time on an evidence state and a stopping rule. Resource-rational metareasoning asks what another step from that state is expected to achieve. This shifts the explanatory question from how difficult a problem is to what progress remains possible during an attempt. A demanding problem can still offer a promising next step, while an unresolved problem can cease to offer a useful direction for search. Humans and models could thus respond to the same difficulty yet continue for different lengths of time because they expect different returns from further work.

\subsection{The value of continued work}
\label{sec:disc-voc}

Resource-rational analysis asks whether further work is worth its cost \citep{lieder2020bbs,gershman2015computational,callaway2022rational}. In a one-step version of the value-of-computation approach \citep{russell1991}, let $s_t$ denote the agent's state after $t$ units of deliberation. Let $V(s_t)$ be the expected utility of stopping with the best response currently available, according to the agent's objective. The marginal value of another unit is
\[
\mathrm{mVOC}(s_t)=
\mathbb{E}[V(s_{t+1})-V(s_t)\mid s_t]-\kappa,
\]
where $\kappa>0$ is the marginal cost of computation. On this criterion, continuation is favoured when the expected improvement from another unit exceeds its cost. Difficulty matters through what it implies about this return. An unresolved problem may still offer a useful next step, or the agent may have exhausted the work it expects to help. A similar level of difficulty can therefore support continuation or stopping, depending on the expected return.

\citet{zhu2026computation} show how limited computation can prevent an agent from fully incorporating available evidence. Their analysis helps explain why unresolved uncertainty need not mean that the information needed to answer is absent. Further processing of what is already available can still be useful. Applied to the present tasks, the question is how much progress an agent expects from continuing to work on the material in the problem.

A quick human failure can follow a judgement that further work has little value. A prolonged failure can arise when the person continues because progress still seems possible, but that work does not yield a solution. Successful attempts may be long for the same reason, with the search eventually succeeding. The H-ARC action records show both brief attempts with few edits and prolonged attempts with substantial grid activity. The human pattern thus includes sustained work as well as early endings. The value of continuing is assessed before the outcome is known, which is why worthwhile work can still end in failure.

For an LRM, a policy that continues to favour further search can produce long traces on attempts that eventually fail. If another reasoning step is still treated as likely to improve the answer, generation can continue through hesitation and returns to earlier material. This explanation locates the human-model difference in how long further search is treated as promising. Whether the additional generation improves the chance of success can be tested by varying the available computation.

The prospect of a better answer is only part of the calculation. Humans and models could expect the same accuracy gain from another step yet value that gain differently. For a person, the value of solving the problem is weighed against the time and effort required to continue. A learned generation policy can be shaped by rewards attached to the output trajectory as well as the final answer, a possibility discussed by \citet{hu2026commentary}. If the policy favours features of a longer response independently of their effect on correctness, continued generation can retain value under that objective after its expected accuracy benefit becomes small. The objective used in $V(s_t)$ and the cost represented by $\kappa$ therefore matter to an explanation of stopping, alongside the expected improvement in the answer.

Visible duration adds a further measurement issue. Another grid operation or another stretch of text records continued activity. How much that activity changes the chance of a correct answer is a separate empirical question. The trace analyses show differences in the composition of failed and successful attempts at comparable lengths. They give us reason to examine what additional generation does, since extending a trace through reconsideration or repetition can have a different effect from adding a step that changes the answer.

The variation across tasks is useful here. On Cortes, successful attempts are longer in both separate subgroup fits, even though the joint comparison retains a human-LRM difference. A control account based on expected progress can accommodate this pattern because continued search can lead to success in either system. H-ARC permits constructive search and substantial grid interaction before a final answer. INTUIT requires physical and social inferences from short vignettes. Cortes presents a binary relational judgement. The available operations and the signs of progress differ across these tasks. Varying the available next steps across otherwise comparable problems would help test whether continuation follows the prospect of further progress.

To estimate the return to further model computation, we propose a randomized reasoning-budget experiment. \citet{hu2026commentary} recommends manipulating trace length while monitoring accuracy. We would randomly assign runs of the same model-item pairs to different token budgets, draw repeated responses under fixed sampling settings, and use a common procedure for eliciting a final answer when reasoning ends or the budget is reached. With $B$ denoting the assigned token budget, the target is
\[
p(\mathrm{correct}\mid\mathrm{do}(B=b),\mathrm{model},\mathrm{item}).
\]
Holding model and item fixed, this intervention would measure how the chance of success changes with the opportunity for further computation. If accuracy rises with the budget, extra computation is useful on average even though some of the longer attempts still fail. Long failed traces could then be part of a policy that improves the overall chance of solving the problem. If accuracy plateaus while traces keep expanding, continued generation contributes little additional success over that range. Measuring hedging and repetition at each budget would help describe how the extra generation is used. The experiment would thus connect the allocation pattern to its consequences for performance.

\subsection{Limitations}
\label{sec:disc-limitations}

The released data support comparisons across attempts. Each model contributes one outcome per item, and the human trials lack participant identifiers. The item-controlled coefficients therefore describe population-level associations between duration and outcome. Repeated model samples and identified human trajectories would permit analysis of individual stopping rules. The proposed budget experiment addresses the effect of additional available computation.

The duration measures record different mixtures of activity. Human reaction time combines several processes \citep{ratcliff2008diffusion,heitz2014sat,bogacz2006}, and reasoning traces imperfectly reflect latent model computation \citep{stechly2025,kambhampati2025anthropomorph,samineni2025}. Logging duration makes the slopes interpretable as within-system relative changes. Their interaction compares these relative outcome differences across systems.

The sample covers six open-weight thinking LRMs from four organisations and one non-thinking baseline. Closed-frontier systems remain outside the comparison. H-ARC and INTUIT provide the clearest opposite-sign subgroup patterns, while Cortes combines a small number of model errors with a different pattern in the separate fits. Extension to other tasks and systems should test which features of the present allocation difference recur and which depend on the search opportunities each task affords.

\section{Conclusion}
\label{sec:conclusion}

The H-ARC comparison shows why agreement about which problems take longer leaves an important part of deliberation unexplained. Model trace length follows the human ordering of problems by duration, yet successful human attempts are longer and failed model attempts have longer traces once item identity is controlled. Human grid actions connect longer attempts to continued task engagement. Failed model traces differ from successful traces at comparable length, with more hedging on H-ARC and more repetition on Cortes.

Separating difficulty registration from deliberation allocation makes deliberative control a distinct target for research on human and model reasoning. Resource-rational analysis directs attention to what another step is expected to achieve and how that gain is valued. The proposed budget experiment would measure when further model computation still improves the chance of success, connecting the observed persistence to its consequences for performance. Understanding similarity in deliberation requires an account of how systems use the opportunity for further thought, as well as a measure of how long they think.

\section*{Acknowledgements}
\addcontentsline{toc}{section}{Acknowledgements}

We thank the authors of the de~Varda et al.\ release for making the
LRM and human reasoning data public, and the H-ARC, INTUIT, and
Cortes data creators for the underlying behavioural datasets.

\section*{Declarations}
\addcontentsline{toc}{section}{Declarations}

\bmhead{Ethics}
This study is a secondary analysis of publicly released,
de-identified behavioural and model-output data. It involved no new
participant recruitment or data collection, and no identifiable
personal data were accessed.

\bmhead{Competing Interests}
The author declares no competing interests.

\bmhead{Data and Code Availability}
All datasets analysed in this study are public. The de~Varda et al.\ release
is at \url{https://osf.io/r3kum/}. H-ARC is accompanied by
\citet{legris2024}, INTUIT by \citet{prunty2025intuit}, and Cortes is
redistributed in the de~Varda deposit. The full analysis pipeline and
intermediate results are openly available at
\href{https://osf.io/jxb4a/}{osf.io/jxb4a}. The repository includes
pinned dependencies, the deterministic random seed
(\texttt{0xA12C}), and a \texttt{Makefile} target that reproduces the
reported results from the public source data.

\FloatBarrier
\bibliographystyle{apalike}
\bibliography{references}

\clearpage
\FloatBarrier
\section*{Supplementary Information}
\addcontentsline{toc}{section}{Supplementary Information}
\setcounter{table}{0}
\setcounter{figure}{0}
\renewcommand{\thetable}{S\arabic{table}}
\renewcommand{\thefigure}{S\arabic{figure}}

\subsection*{SI-1. Single-LRM item fixed-effects interactions on H-ARC}

Table~\ref{tab:itemfe-perlrm-si} reports the combined item-FE
interaction after refitting the human sample with each thinking LRM
separately. The four H-ARC model samples with at least $298$
observations all return negative interactions at $p<10^{-4}$. The gpt-oss-20b estimate is treated
descriptively because only $119$ final grids are parseable, and the
Qwen3-235B-Thinking estimate is descriptive because only four of its
$43$ released H-ARC trials are wrong. The corresponding per-agent
Cohen's $d$ values for the estimable cells are shown in
Figure~\ref{fig:harc-primary}a.

\begin{table}[!ht]
\centering
\small
\begin{tabular}{lrrrrr}
\toprule
Model & $n_\text{LRM}$ & $n_\text{wrong}$ & $\beta_{\mathrm{LRM}}$ & $95\%$ CI & $p$ \\
\midrule
Qwen-QwQ-32B          & 400 & 380 & $-0.831$ & $[-1.22,-0.44]$ & $2.9\times10^{-5}$ \\
gpt-oss-20b\textsuperscript{\textdagger}           & 119 &  91 & $-0.815$ & $[-1.24,-0.39]$ & $1.9\times10^{-4}$ \\
gpt-oss-120b          & 389 & 322 & $-0.808$ & $[-1.05,-0.57]$ & $5.0\times10^{-11}$ \\
GLM-4.5-Air-FP8       & 388 & 356 & $-0.731$ & $[-1.00,-0.46]$ & $8.3\times10^{-8}$ \\
DeepSeek-R1           & 298 & 237 & $-0.584$ & $[-0.75,-0.42]$ & $1.3\times10^{-12}$ \\
Qwen3-235B-Thinking\textsuperscript{\textdagger}   &  43 &   4 & $-0.323$ & $[-0.56,-0.09]$ & $6.4\times10^{-3}$ \\
\botrule
\end{tabular}
\caption{\textbf{H-ARC single-LRM item-FE interactions}
(\textit{The allocation gap on H-ARC}).
$\beta_{\mathrm{LRM}}$ is the within-item correctness-slope
difference between the LRM and humans, with item fixed effects and cluster-robust SE by item.
\textsuperscript{\textdagger}\,The dagger marks descriptive estimates.
gpt-oss-20b has $n=119$ parseable final outputs. Qwen3-235B-Thinking is
identified from four wrong trials.}
\label{tab:itemfe-perlrm-si}
\end{table}

\subsection*{SI-2. H-ARC robustness and alternative specifications}
\label{sec:si-harc-robustness}

\textit{Within--between decomposition.}
The Mundlak parameterisation separates the within-item correctness
slope $\beta_W$ from the between-item component $\beta_M$. The
between-item slope is $\beta_W+\beta_M$.

\begin{table}[!ht]
\centering
\small
\begin{tabular}{lrrr}
\toprule
Subgroup & Within-item $\beta_W$ & Between-item $\beta_W{+}\beta_M$ & Mundlak $\beta_M$ ($p$) \\
\midrule
Humans only        & $+0.241$ & $-0.535$ & $-0.777$ ($5\times10^{-22}$) \\
LRMs (pooled)      & $-0.234$ & $-1.214$ & $-0.980$ ($2\times10^{-17}$) \\
\multicolumn{4}{l}{Combined frame, $\beta_{\mathrm{LRM}}=-0.793$ ($p=8\times10^{-28}$)} \\
\botrule
\end{tabular}
\caption{\textbf{Mundlak within--between decomposition on H-ARC}
(\textit{The allocation gap on H-ARC}). The
combined-frame dissociation under Mundlak ($-0.79$) is larger in
magnitude than the primary item-FE estimand ($-0.66$).}
\label{tab:itemfe-mundlak-si}
\end{table}

\textit{Aggregation.}
The public release contains multiple human trials per item and omits
participant identifiers. We
collapse each human item$\times$correctness cell to its
mean $\log t$, once weighted by cell size and once unweighted, while
leaving each LRM model$\times$item row unchanged. Both aggregated
fits reproduce the negative interaction from the trial-level model
(Table~\ref{tab:aggcell}).

\begin{table}[!ht]
\centering
\small
\begin{tabular}{lrrrrl}
\toprule
Specification & $n_{\text{obs}}$ & $\beta_{\mathrm{LRM}}$ & SE & $p$ & $95\%$ CI \\
\midrule
Trial-level (primary) & $5{,}728$ & $-0.655$ & $0.078$ & $4\times10^{-17}$ & $[-0.81,-0.50]$ \\
Cell-aggregated, weighted & $2{,}417$ & $-0.655$ & $0.082$ & $2\times10^{-15}$ & $[-0.82,-0.49]$ \\
Cell-aggregated, unweighted & $2{,}417$ & $-0.606$ & $0.070$ & $4\times10^{-18}$ & $[-0.74,-0.47]$ \\
\botrule
\end{tabular}
\caption{\textbf{Aggregated-cell robustness on H-ARC.} Human trials
are collapsed to (item$\times$correctness) cells. LRM rows are left
at one observation per (model$\times$item). The combined item-FE
dissociation regression is then re-estimated. Weighted by cell size
or unweighted, the primary interaction coefficient is essentially
unchanged from the trial-level estimate.}
\label{tab:aggcell}
\end{table}

\textit{Alternative specifications and identification.}
The item fixed-effects model with item-clustered standard errors is
the primary difficulty-controlled specification. Table~\ref{tab:harc-spec-si}
reports four sensitivity analyses using random effects or alternative
clustering. They differ in how between-item information and agent-level
variation enter the fit, so their coefficient magnitudes are not
interpreted as directly interchangeable. The agent-clustered OLS fit
has only seven clusters and is included as a finite-cluster
sensitivity analysis.

In the pooled LRM item-FE regression, the outcome slope is identified
by variation across models attempting the same item. A single-LRM combined
fit is identified by variation across the human trials and that model
on the same item. The released data contain one outcome per
model$\times$item pair, so none of these regressions estimates a
repeated stochastic within-item slope for one model.

\begin{table}[!ht]
\centering
\small
\begin{tabular}{lrrrr}
\toprule
Specification & Interaction $\beta$ & SE & $z/t$ & $p$ \\
\midrule
Item random intercept (mixed-effects anchor) & $-0.660$ & $0.054$ & $-12.31$ & $<.001$ \\
Crossed random agent + item & $-0.940$ & $0.058$ & $-16.24$ & $<.001$ \\
OLS, cluster SE by agent & $-0.862$ & $0.089$ & $-9.67$  & $<.001$ \\
OLS, cluster SE by item  & $-0.862$ & $0.073$ & $-11.77$ & $<.001$ \\
\midrule
\textbf{Item fixed effects (primary estimand)} & $\mathbf{-0.655}$ & $\mathbf{0.078}$ & $\mathbf{-8.42}$ & $\mathbf{<.001}$ \\
\botrule
\end{tabular}
\caption{\textbf{H-ARC dissociation interaction across
specifications.} Centred correctness $\times$ LRM interaction.
The bold row is the primary difficulty-controlled estimand
(within-item identification, cluster-robust SE by item). The four
upper rows are mixed-effects and cluster-robust sensitivity analyses
that return the same interaction sign without imposing strict
within-item identification. The agent-clustered OLS row is a
finite-cluster sensitivity check with only seven clusters, the
human population plus the six thinking LRMs. All five
specifications agree in sign.}
\label{tab:harc-spec-si}
\end{table}

\subsection*{SI-3. LRM data provenance and scoring}
\label{sec:si-provenance}

The released frame (\citealp{devarda2025}) contains each model's
prompt, full output, final-answer fields, and token counts. For
thinking LRMs we use \texttt{reasoning\_token\_length}, counted in
the model's native tokenizer. For DeepSeek-V3 we use
\texttt{total\_output\_tokens}. Models are not retokenized onto a
common scale, and raw token counts are not compared across models.
Final-answer correctness follows the parser used in the de Varda et
al. analysis, augmented for H-ARC grid outputs by a
regular-expression match against \texttt{Correct\_answer}. Full
attempted, parseable, correct, and wrong counts are provided in the
OSF deposit.

\textit{gpt-oss-20b.} Only $119$ of $400$ final grids are parseable.
The remaining outputs have no correctness label, so this cell is
reported descriptively. \textit{Qwen3-235B-Thinking.} Only $43$ H-ARC
items appear in the released frame and four are wrong, so its H-ARC
single-model estimates are also treated descriptively.

\subsection*{SI-4. Multiple-comparison treatment}
\label{sec:si-multcomp}

The reported $p$-values are unadjusted. Treating the three combined paradigm interactions as one family,
all three interactions remain significant under Bonferroni and Holm correction. Among the eight
trace-content feature$\times$paradigm tests in
Table~\ref{tab:trace-content}, the three effects used in the main-text
interpretation also survive Bonferroni correction. The H-ARC type--token-ratio effect has $p\approx.0063065$, just above the Bonferroni threshold of $.05/8=.00625$. Secondary cells not used as primary
inferential evidence are treated descriptively in the text.

\subsection*{SI-5. Cross-paradigm item-FE regressions and single-LRM interactions}
\label{sec:si-paradigms}

Table~\ref{tab:itemfe-paradigm} reports the human-only, pooled-LRM,
and combined item-FE regressions for H-ARC, INTUIT, and Cortes.
Arithmetic is omitted because every thinking LRM solves every item.
Separate subgroup fits estimate item effects within each group.
The combined regression imposes shared item effects and estimates both
outcome slopes jointly on items represented in both groups. Its
interaction is the jointly estimated LRM slope minus the human slope.
On Cortes, the human-only regression uses $20{,}052$ trials on $86$
items, the LRM-only regression uses $348$ trials on $58$ items, and
the combined regression uses $13{,}888$ observations on the $58$
shared items. The combined human slope is $+0.471$ and the implied
LRM slope is $-0.497$, yielding the interaction of $-0.968$.
Each Cortes model contributes only $5$--$8$ wrong trials.

\begin{table}[!ht]
\centering
\small
\begin{tabular}{lrrrrrrrr}
\toprule
& & & \multicolumn{2}{c}{Humans only} & \multicolumn{2}{c}{LRMs pooled} & \multicolumn{2}{c}{Combined dissoc.} \\
\cmidrule(lr){4-5}\cmidrule(lr){6-7}\cmidrule(lr){8-9}
Paradigm  & $n_\text{H}$ & $n_\text{LRM}$ & $\bcorrect{}$ & $p$ & $\bcorrect{}$ & $p$ & $\beta_{\mathrm{LRM}}$ & $p$ \\
\midrule
H-ARC      & 4{,}091  & 1{,}637 & $+0.241$ & $<.001$ & $-0.271$            & $<.001$ & $-0.655$ & $<.001$ \\
INTUIT     & 1{,}536  &     864 & $+0.102$ & $.087$  & $-0.331$            & $.013$  & $-0.410$ & $<.001$  \\
Cortes     & 20{,}052 &     348 & $+0.418$ & $<.001$ & $\mathbf{+0.310}$   & $.001$  & $-0.968$ & $<.001$ \\
\botrule
\end{tabular}
\caption{\textbf{Cross-paradigm item-FE regressions.} Three
within-item regressions per paradigm, with cluster-robust SE by
item. The bold entry marks the positive LRM slope in a separate subgroup
fit. Sample sizes in the first two columns refer to the subgroup
fits, not necessarily the combined regression.}
\label{tab:itemfe-paradigm}
\end{table}

Table~\ref{tab:perlrm-cross} reports the corresponding combined
regression after pairing the human sample with one LRM at a time.
Every model--paradigm interaction is negative. The limited H-ARC
cells for gpt-oss-20b and Qwen3-235B-Thinking remain descriptive for
the reasons given in SI-1 and SI-3.

\begin{table}[!ht]
\centering
\small
\begin{tabular}{lrrr}
\toprule
Model & H-ARC & INTUIT & Cortes \\
\midrule
Qwen-QwQ-32B          & $-0.83$ & $-0.73$ & $-1.51$ \\
gpt-oss-120b          & $-0.80$ & $-0.27$ & $-1.20$ \\
GLM-4.5-Air-FP8       & $-0.73$ & $-0.29$ & $-0.94$ \\
gpt-oss-20b           & $-0.78$ & $-0.69$ & $-0.77$ \\
DeepSeek-R1           & $-0.57$ & $-0.14$ & $-0.73$ \\
Qwen3-235B-Thinking   & $-0.27$ & $-0.18$ & $-0.77$ \\
\botrule
\end{tabular}
\caption{\textbf{Single-LRM interaction $\beta_{\mathrm{LRM}}$ across
paradigms.} Combined regression refit on humans plus each single
LRM, with item fixed effects and cluster-robust SE by item. Each fit uses items shared by the human sample and the particular
LRM. On H-ARC this restriction is applied separately for each model.
Table~\ref{tab:itemfe-perlrm-si} retains the full human sample,
including items absent from the paired model's data. The resulting
differences are largest for the two limited H-ARC cells.}
\label{tab:perlrm-cross}
\end{table}

\subsection*{SI-6. Trace-content regressions}
\label{sec:si-trace}

Table~\ref{tab:trace-content} reports all eight length-controlled
feature$\times$paradigm coefficients specified in the
\textit{Trace-content analysis} section of Methods. The three rows emphasised
in the main text are shown in bold. INTUIT reasoning traces are
restricted in the public release and are not included in this
analysis.

\begin{table}[!ht]
\centering
\small
\begin{tabular}{llrrrl}
\toprule
Paradigm & Feature & $\beta$ & SE & $p$ & $95\%$ CI \\
\midrule
H-ARC & Hedge-marker density / 1k chars & $\mathbf{-0.435}$ & $0.128$ & $7\times10^{-4}$ & $[-0.69,-0.18]$ \\
H-ARC & Self-correction density  & $-0.0018$ & $0.0012$ & $.15$ & $[-0.004,+0.001]$ \\
H-ARC & 5-gram repetition rate   & $+0.016$ & $0.009$  & $.09$ & $[-0.002,+0.034]$ \\
H-ARC & Type--token ratio        & $-0.004$ & $0.001$  & $.006$ & $[-0.007,-0.001]$ \\
Cortes & Hedge-marker density / 1k chars & $-0.817$ & $0.834$ & $.33$ & $[-2.45,+0.82]$ \\
Cortes & Self-correction density & $-0.0021$ & $0.0015$ & $.16$ & $[-0.005,+0.001]$ \\
Cortes & 5-gram repetition rate  & $\mathbf{-0.074}$ & $0.014$ & $7\times10^{-8}$ & $[-0.10,-0.05]$ \\
Cortes & Type--token ratio       & $\mathbf{+0.041}$ & $0.012$ & $10^{-3}$ & $[+0.02,+0.06]$ \\
\botrule
\end{tabular}
\caption{\textbf{Trace-content regressions, length-controlled.} For
each feature, $\beta$ is the coefficient on correctness in
$\text{feature}\sim\text{correct}+\log w+C(\text{agent})+C(\text{item})$
with cluster-robust SE by item. A negative $\beta$ on the
hedge-marker density or repetition row means the feature is
\textit{higher} on wrong trials. Bold rows are the emphasised
content asymmetries on each paradigm. INTUIT traces are restricted
in the public release.}
\label{tab:trace-content}
\end{table}

\subsection*{SI-7. Additional robustness and uncertainty figures}
\label{sec:si-robust}

\begin{figure}[!ht]
  \centering
  \includegraphics[width=\textwidth]{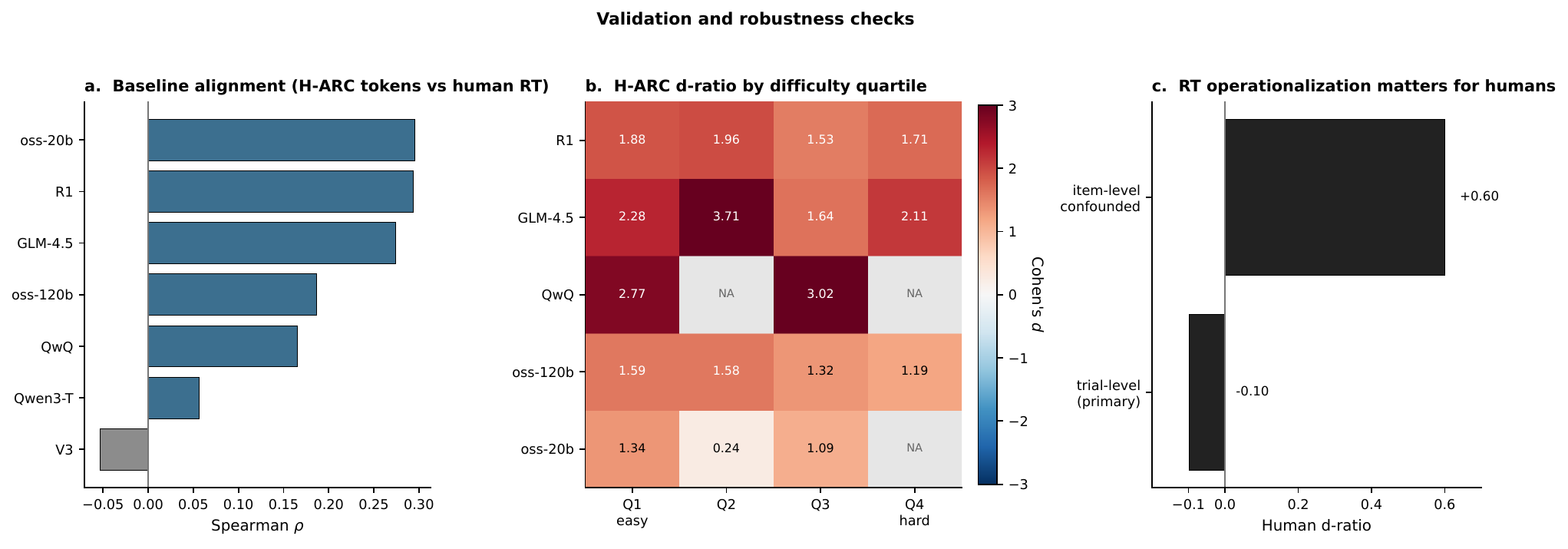}
  \caption{\textbf{Robustness checks for the H-ARC allocation gap.}
  (a)~Cross-item Spearman correlation between LRM reasoning-token
  length and human reaction time on H-ARC. DeepSeek-V3
  is the non-thinking baseline and shows no positive correlation.
  (b)~LRM d-ratios by difficulty quartile. Cells marked
  ``NA'' have fewer than five right or five wrong trials, so no d-ratio is reported.
  (c)~Human d-ratio under the trial-level (primary) and
  item-level operationalisations. The item-level contrast groups items by released accuracy at or below $50\%$ versus above it.}
  \label{fig:robustness}
\end{figure}

\begin{figure}[!ht]
  \centering
  \includegraphics[width=0.82\textwidth]{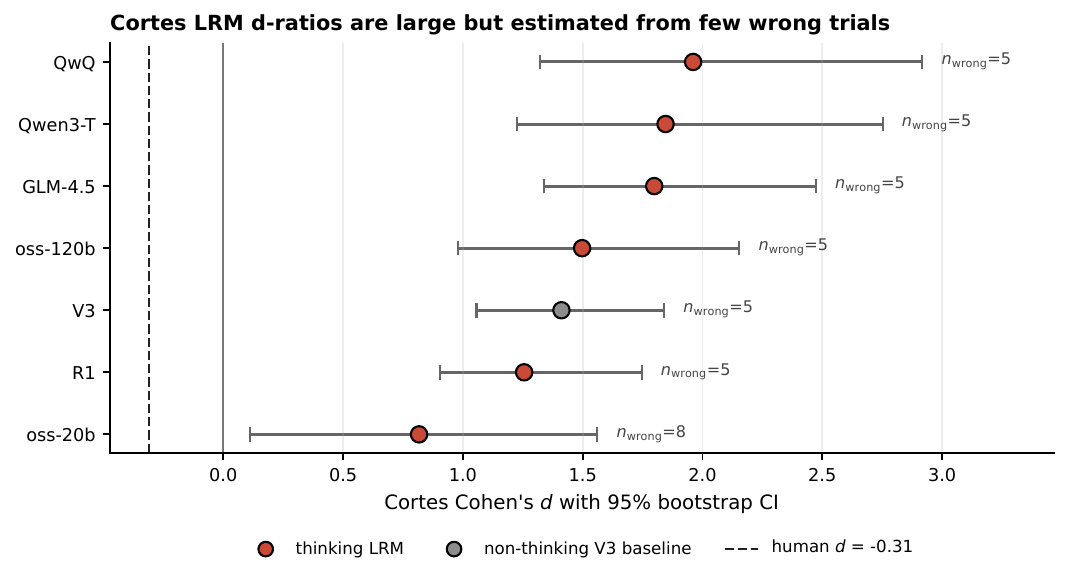}
  \caption{\textbf{Cortes bootstrap uncertainty.} Points are Cortes
  d-ratios per agent. Horizontal bars are $95\%$ bootstrap
  confidence intervals from $10{,}000$ resamples. Per-LRM
  wrong-trial counts are annotated on the right of each interval.
  The dashed line marks the human Cortes d-ratio. The non-thinking
  DeepSeek-V3 baseline is included for reference.}
  \label{fig:cortes-ci}
\end{figure}

Figure~\ref{fig:robustness} collects additional H-ARC robustness
checks. Figure~\ref{fig:cortes-ci} reports Cortes per-agent
nonparametric bootstrap intervals \citep{efron1993bootstrap}.

\end{document}